\documentclass[journal]{IEEEtran}

\usepackage{times}
\usepackage{epsfig}
\usepackage{graphicx}
\usepackage{amsmath}
\usepackage{amssymb}
\usepackage[linesnumbered,ruled]{algorithm2e}
\usepackage{multirow}
\usepackage{xcolor}
\usepackage[normalem]{ulem}
\usepackage{slashbox}
\usepackage{caption}
\usepackage{subcaption}
\usepackage{dblfloatfix}

\usepackage{tabularx}
\usepackage{array}
\usepackage{enumitem}

\makeatletter
\newcommand{\thickhline}{%
    \noalign {\ifnum 0=`}\fi \hrule height 1pt
    \futurelet \reserved@a \@xhline
}
\newcolumntype{"}{@{\hskip\tabcolsep\vrule width 1pt\hskip\tabcolsep}}
\makeatother

\makeatletter
\newcommand{\nosemic}{\renewcommand{\@endalgocfline}{\relax}}
\newcommand{\dosemic}{\renewcommand{\@endalgocfline}{\algocf@endline}}
\let\oldnl\nl
\newcommand{\nonl}{\renewcommand{\nl}{\let\nl\oldnl}}
\makeatother

\ifCLASSINFOpdf
\else
\fi
\begin{document}
%


 \title{A Semi-supervised Spatial Spectral Regularized Manifold Local Scaling Cut With HGF for Dimensionality Reduction of Hyperspectral Images}
%
%
%

\author{Ramanarayan~Mohanty,~\IEEEmembership{Student Member,~IEEE,}
        S~L~Happy,~\IEEEmembership{Member,~IEEE,}
        and~Aurobinda~Routray,~\IEEEmembership{Member,~IEEE}
\thanks{R. Mohanty is with the Advanced Technology Development Centre, Indian Institute of Technology, Kharagpur,
West Bengal, 721302 India e-mail: ramanarayan@iitkgp.ac.in.}
\thanks{S L Happy and A. Routray are with Department of Electrical Engineering, Indian Institute of Technology, Kharagpur.}}
\maketitle

\begin{abstract}

Hyperspectral images (HSI) contain a wealth of information over hundreds of contiguous spectral bands, making it possible to classify materials through subtle spectral discrepancies. However, the classification of this rich spectral information is accompanied by the challenges like high dimensionality, singularity, limited training samples, lack of labeled data samples, heteroscedasticity and nonlinearity. To address these challenges, we propose a semi-supervised graph based dimensionality reduction method  named `semi-supervised spatial spectral regularized manifold local scaling cut' (S3RMLSC). The underlying idea of the proposed method is to exploit the limited labeled information from both the spectral and spatial domains along with the abundant unlabeled samples to facilitate the classification task by retaining the original distribution of the data. In S3RMLSC, a hierarchical guided filter (HGF) is initially used to smoothen the pixels of the HSI data to preserve the spatial pixel consistency. This step is followed by the construction of linear patches from the nonlinear manifold by using the maximal linear patch (MLP) criterion. Then the inter-patch and intra-patch dissimilarity matrices are constructed in both spectral and spatial domains by regularized manifold local scaling cut (RMLSC) and neighboring pixel manifold local scaling cut (NPMLSC) respectively. 
Finally, we obtain the projection matrix by optimizing the updated semi-supervised spatial-spectral between-patch and total-patch dissimilarity. The effectiveness of the proposed DR algorithm is illustrated with publicly available real-world HSI datasets.

\end{abstract}

\begin{IEEEkeywords}
Dimensionality reduction, hyperspectral image, manifold local scaling cut, neighboring pixel manifold local scaling cut, regularized manifold local scaling cut, semi-supervised spatial-spectral regularized manifold local scaling cut.
\end{IEEEkeywords}

%
\IEEEpeerreviewmaketitle

\section{Introduction}
\IEEEPARstart {H}{yperspectral} remote sensing images (HSI) with high spectral and spatial resolutions capture the inherent physical and chemical properties of the land cover. Therefore, HSI data analysis finds potential applications in environmental research, geological surveys, mineral identification, agriculture monitoring, etc. However, the availability of limited train data with a large number of spectral bands, make these applications very challenging. Generally the variation in sun-canopy-sensor geometry, the multipath scattering of light and non-homogeneous composition of pixels make the acquired HSI data modeling nonlinear \cite{ghamisi2017advances}. 
Handling these complex and high dimensional redundant nonlinear data is a major challenge in HSI data analysis. To mitigate this challenge in HSI data analysis, 
 an effective dimensionality reduction (DR) method is essential before training the classifiers. In this paper, we have made an effort to address this challenge from the manifold learning point of view.

Assuming the real world high dimensional data possess few degrees of freedom \cite{zheng2009statistical}, manifold learning helps in recovering compact, meaningful low dimensional structures from the complex high dimensional data for subsequent processing, such as classifications and visualizations \cite{lunga2014manifold}. This projects the higher dimensional data into lower dimensional space, while preserving their underlying geometrical structure \cite{lin2008riemannian}. Several state-of-the-art techniques for DR utilize manifold learning, such as local linear embedding (LLE) \cite{roweis2000nonlinear}, isometric feature mapping (ISOMAP) \cite{tenenbaum2000global}, etc. These methods follow unsupervised mode to model the data using a single manifold. However, the DR methods are mainly classified into three major categories unsupervised, supervised and semi-supervised ones. The unsupervised DR methods project the data to lower dimensional space without using any label information. Several state-of-the-art unsupervised DR techniques include principal component analysis (PCA) \cite{martinez2001pca}, LLE \cite{roweis2000nonlinear}, \cite{han2005nonlinear}, ISOMAP \cite{tenenbaum2000global}, Laplacian eigenmap (LE) \cite{belkin2001laplacian}, etc. However, their implicit nonlinear mapping technique forbids them to directly apply to new test samples. This limits the application of these methods in the classification task. Apart from that, these unsupervised manifold learning methods search $k$-nearest neighbors (Knn) of the given point from different classes, whereas their supervised counterparts only identify the neighbors that are of the same class of that given point. Hence, it makes the supervised manifold learning approach more favorable towards the classification of HSI data \cite{geng2005supervised}, \cite{ma2010local}. 

The supervised DR approaches use the label information to learn the discriminative projections. This includes, linear discriminant analysis (LDA) \cite{belhumeur1997eigenfaces}, \cite{bandos2009classification}, scaling cut (SC) \cite{zhang2009local}, local scaling cut (LSC) \cite{zhang2015scaling}, \cite{zhang2013semisupervised}, \cite{zhang2009local}, linear discriminant embedding (LDE) \cite{chen2005local}, local fisher discriminant analysis (LFDA) \cite{sugiyama2007dimensionality}, \cite{li2012locality}, nonparametric weighted feature extraction (NWFE) \cite{kuo2004nonparametric} and so on. The LDA seeks discriminative projection by maximizing between-class scatter and minimizing within-class scatter by assuming the class distribution as unimodal Gaussian distribution with equal covariance. Therefore, LDA fails to handle real world heteroscedastic and multimodal data. SC \cite{zhang2009local} and LSC \cite{zhang2015scaling} address these issue by constructing pairwise dissimilarity matrix among the samples. LDE extend the LDA by performing the local discriminant in a graph embedding framework. LFDA approach fuse the discriminative property of LDA with the local preserving capability of locality preserving projection (LPP) \cite{he2005face}. The NWFE method extends the LDA method by adding a nonparametric scatter matrices with training samples. 

The aforementioned DR methods mostly focus on the spectral based approaches. They use the spectral domain Euclidean distance to compute the similarity measure. The spectral-domain methods possess several limitations, such as: 1) Relatively large spectral bands with respect to small training samples create a singularity in the sample covariance matrix that leads to ill-posed problems in classification \cite{he2018recent}. 2) The limited availability of labeled data samples for supervised learning is a major bottleneck in HSI data classification. It is also an expensive task in terms of time and efforts to label all the acquired data for classification purpose. 3) As the HSI data class is distributed to multiple subregions, two HSI data samples with small spectral distance measures may have large spatial distances or may be belong to different class (e.g. The concrete roof top of a house and concrete road may have similar spectral similarity measure but they belong to different class). This implies that only spectral similarity measure is not sufficient for HSI data classification task. Hence, the spectral similarity measure without considering the spatial inter-pixel correlation sometimes leads to under-classification or over-classification \cite{zhou2015dimension}. Therefore, use of spectral information alone results in unsatisfactory performances.   
To mitigate the challenge of limited training data, several DR algorithms adopt the semi-supervised approach by incorporating the unlabeled data samples with the labeled training data samples. Several attempts have been made for semi-supervised classification of HSI data, such as semi-supervised discriminant analysis (SDA) \cite{cai2007semi}, semi-supervised local fisher discriminant analysis (SELF) \cite{sugiyama2010semi}, semi-supervised local discriminant analysis (SELD) \cite{liao2013semisupervised}, generalized semi-supervised local discriminant analysis (GSELD) \cite{liao2012classification}, semi-supervised local scaling cut (SSLSC) \cite{zhang2013semisupervised}. The SDA is the semi-supervised version of the LDA and it faces the similar problem as LDA faces. Similarly, SELF, SELD, GSELD and SSLSC are the semi-supervised versions of the LFDA, NPE and LSC algorithms. The SELF is based on LDA, when the availability of labeled samples are less, LDA performs poorly due to over-fitting of the data. The SELD method has addressed the issues faced by SELF. Still, it overlooks the non-linear property of the labeled data in deriving the scatter matrix for the whole class. The GSELD method is the extended version of SELD with the tunable parameters. Similarly, SSLSC only considers the $k$-nearest elements to form the dissimilarity matrix without considering non-linear property of the classes. However, this is not sufficient to cope with other mentioned limitations. Hence, supervised DR methods consider both the spatial as well as spectral information to perform the similarity measure.  


The spatial contexture information boosts the discrimination ability of the spectral based information to improve HSI data classification performance. Therefore, spectral-spatial based methods gain considerable attention in HSI feature extraction and classification tasks \cite{he2018recent,zhou2015dimension,xia2015spectral,sun2015supervised}. In \cite{zhong2015discriminant}, Zhong. et. al proposed a tensorial extension of LDA to extract spatial spectral features of HSI, which obeys the Gaussian distribution with equal variance. The spatial methods include two major category i) spatial filtering ii) spatial feature extraction. The spatial filtering method is a preprocessing approach to classification, while spatial feature extraction incorporates the spatial information to improve the class discrimination. For example, multiple spectral and spatial features at both pixel and object levels are combined in \cite{huang2013svm} to construct an ensemble support vector machine (SVM) for direct classification of the data. 

These above mentioned methods either focus more on the data distribution problem or pay more attention to the problem of learning discriminant function by overlooking the nonlinearity property, intrinsic geometry and proper data projection direction. In this paper, we have proposed an approach for solving the above issues, that other techniques have overlooked. To achieve the objective, we first represent the whole dataset as the union of several local linear patches followed by the application of the local scaling-cut to find the optimal projection matrix of the data in both spectral and spatial domains. Then we extend it to semi-supervised method by exploiting the unlabeled data samples and named it as semi-supervised spatial-spectral regularized manifold local scaling cut (S3RMLSC). The brief overview of the proposed method is discussed below.

  Here, we propose the manifold based method for reducing the dimensions of the HSIs on the basis of their geometry and nonlinearity property. The proposed semi-supervised spatial-spectral regularized manifold local scaling cut (S3RMLSC) method is derived in six small steps. The initial step of the method consists of preprocessing of the data by a spatial hierarchical guided filter (HGF) \cite{pan2017hierarchical} to enhance the pixel consistency by performing the edge aware noise smoothening. Next in the second step, we generate the local patches from the preprocessed data by selecting data points of the local neighborhood within the class in hierarchical manner. This results in construction of multiple non overlapping linear patches from a single class. In the third step, we propose the spectral domain manifold scaling cut (MLSC) by constructing the inter-patch and intra-patch dissimilarity matrices. These inter-patch dissimilarity matrix is constructed between two nearest patches of different classes only. Then, we add a regularizer with the spectral domain MLSC to improve the classification performance by enhancing data diversity and stability and formulated a regularized MLSC (RMLSC). However, spectral information is not sufficient to achieve better classification accuracy. Hence, in the fourth step, we propose a graph based spatial segmentation technique of the patches, called neighboring pixel MLSC (NPMLSC). This NPMLSC method constructs the between-patch and within-patch dissimilarity matrix among the spectrally closest patches. After getting the dissimilarity matrices by spectral RMLSC and spatial NPMLSC, we fuse both the dissimilarity matrices to achieve the new dissimilarity matrix for spectral-spatial MLSC (SSRMLSC) in step five. Finally, in the last step, we incorporate the unlabeled data with the labeled data to formulate the semi-supervised SSRMLSC (S3RMLSC) method.
  Irrespective of the distribution and the modality of the data, these local patches in the manifold are better separated, and meanwhile, the intrinsic geometry of data is well preserved to maintain the within patch compactness. That assures the reliable classification of the new testing data in the projected embedding space.

\begin{figure*}[!htp]
	\begin{center}
		\includegraphics [width=0.75\textwidth] {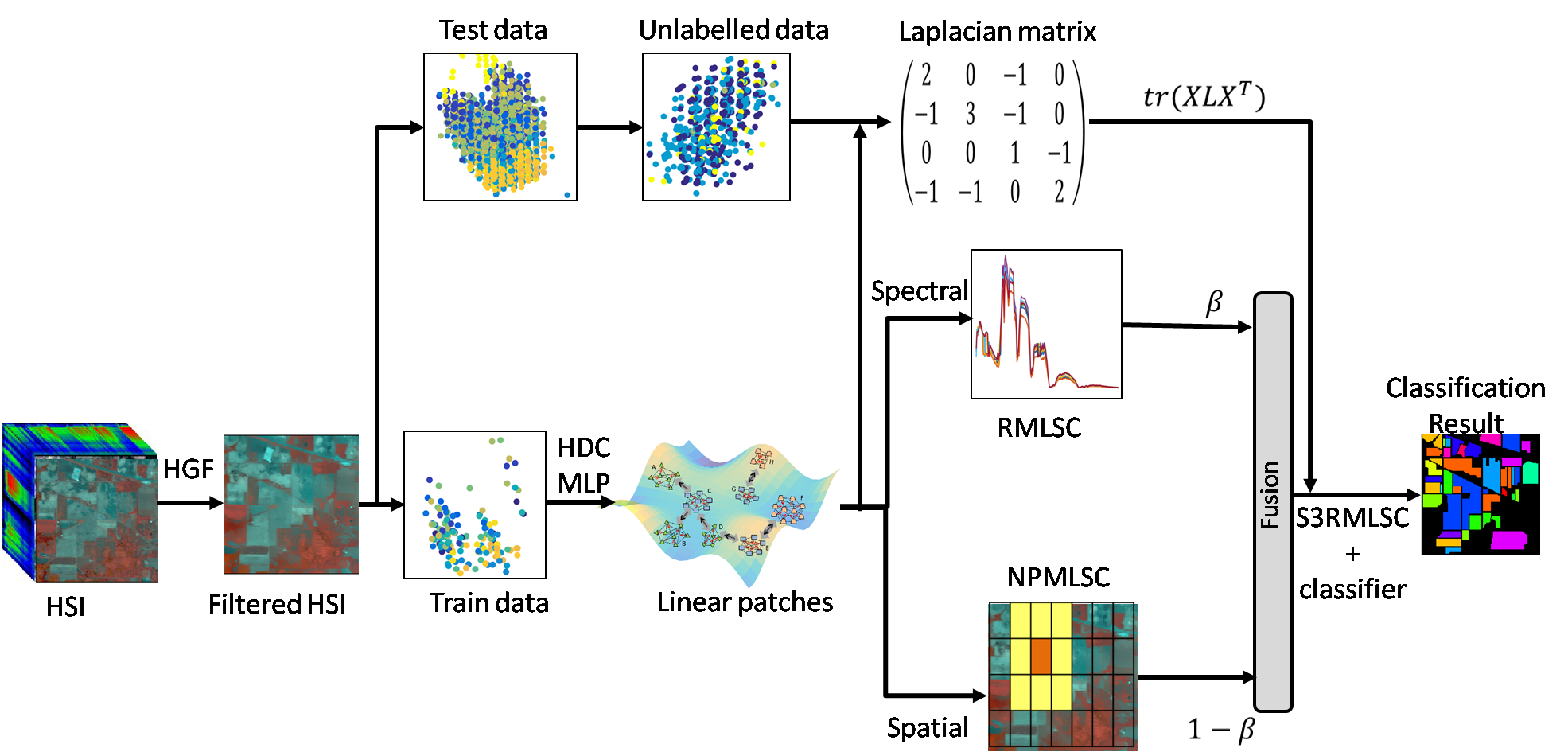}
	\end{center}
    \caption{Complete workflow diagram of the proposed approach.}
	\label{fig:workflow}
\end{figure*}

In summary, the method proposed in this paper addresses several issues like high dimensionality, singularity, insufficient labeled data samples, multimodality, heteroscedasticity while preserving local geometry of the  data. The major characteristics of this work are enumerated as follows:
\begin{itemize}
	\item The spatial filter enhances the robustness towards the noisy points by preserving the neighboring pixel consistency.
	\item The spectral information reveal the nonlinear manifold properties of the HSI data and gather the neighboring data samples from the manifold that spans the same class and lie on the linear patch.
    \item The spatial inter-pixel correlation among the elements of the nearest patches enhances the class discrimination of the objects having similar spectral signature but belong to different class.  
    \item The regularization strategy used in the spectral domain graph cut overcomes the singularity issue that might arise due to small size samples. Moreover, the penalty term reveals the data diversity and enhances the data stability.
    \item SSRMLSC fuses both local label neighborhood as well as the local pixel neighborhood relation of the patches to achieve better projection and high classification accuracy. 
	\item S3RMLSC incorporates spectral-spatial information from the labeled train data with the randomly selected unlabeled data samples from the test data for further improvement of the classification performance. 
\end{itemize}  

The complete work-flow diagram of the proposed S3RMLSC based HSI classification system is shown in Fig.~\ref{fig:workflow}. 

\section{Semi-supervised Spatial Spectral Regularized Manifold Local Scaling Cut } \label{sec:sdmlsc}

\subsection{Problem formulation}

Let's assume that the data samples of the dataset lie on a nonlinear manifold $\mathcal{M}$. 
In semi-supervised learning, a few unlabeled samples from test set are used during training. Suppose, $L+U$ is the total number of training data containing $L$ number of labeled and $U$ unlabeled samples. Therefore, we can represent the training dataset as $X = {x_i}|_{i=1}^{L+U} = (X_L, X_U); x_i  \in R^\mathbb{D}$, where $X_L = {x_i}|_{i=1}^L$ is the labeled training dataset with label $y_i$ and $X_U$ is the unlabeled training data collected from the test dataset. The test dataset is $X^{test}$ and $X_U \subset X^{test}$. Here the number of distinct classes is $K$, i.e. $y_i \in \{1,2, ...,K\}$.  
We represent $\mathcal{M}$ as the union of several linear patches such that $\mathcal{M} = \{S_{1}, S_{2}, ... , S_{n}\}$, where $n$ denotes the number of linearized patches.



Our objective is to project the higher dimensional feature space to a lower one ($R^d$) by considering the original distribution of samples in the patch-wise locality of the spectral domain manifold while utilizing the information from spatial neighborhood pixel structure to boost the system performance. These projection directions are obtained by maximizing the between-patch separability and minimizing the within-patch distances to enhance compactness of the local patches with varied class labels. First, the HSI data are preprocessed spatially by HGF \cite{pan2017hierarchical}. Then, the manifold is learned considering the spatial-spectral local patch discrimination to distinguish manifold boundaries efficiently.


\subsection{Spatial Hierarchical Guided Filter} \label{sec:HGF}
The HGF \cite{pan2017hierarchical} is a hierarchical extension of edge preserving guided filter (GF) \cite{kang2014spectral} which is used for edge-aware noise removal. 
It is based on an assumption of local linear model, i.e., the filter output $F$ is a linear transformation of the guidance image $I$ in a squared window $w_k$ of size $r \times r$ centered at the pixel $k$:
\begin{equation} \label{eq:gf_LT}
f_i = {a_k}{I_i} + {b_k} \,\,\,\,\, \forall i \in {w_k}
\end{equation}
where ${a_k}$ and ${b_k}$ are some linear coefficients for $w_k$. The assumption of this model ensures that ${\nabla}f \approx a{\nabla}I$, i.e., the filter output $F$ has an edge if the guidance image $I$ has an edge at that location. Given the input image $P$, the linear coefficients ${a_k}$ and ${b_k}$ are determined by minimizing the energy function:
\begin{equation} \label{eq:gf_energy}
E({a_k},{b_k}) = \sum_{i \in w_k}{(({a_k}{I_i} + {b_k}-P_i)^2+ \epsilon {a_k}^2)}
\end{equation}
Here $\epsilon$ is the regularization parameter that controls the degree of blurring of the guided image. 

In this filtering step, we initially obtain the guidance image $I$ by taking PCA of the input image $P$ and the obtained first principal component is selected as the gray scale single channel guidance image, so that maximum reconstruction is possible. The given input HSI dataset is represented as $X = \{B_1, B_2, ..., B_D\}$, where the input image $P = X$, $B_i$ is the $i$th band and $D$ is the total number of bands. The principal components of the HSI data is derived as $[pc_1, pc_2, pc_3, ...,pc_D] = PCA (X)$, and the constructed guidance image $I = [pc_1]$. Then, using Eq.~\ref{eq:gf_LT} and \ref{eq:gf_energy}, we determine the filtered output of each band of input image $P$ and generate the new filtered image $F = [f_1, f_2, ... , f_D]$ with dimensions same as data $P$. In this hierarchical model the output image $F$  of the current hierarchy is utilized as the input to the next hierarchy. This filter captures different small and large homogeneous spatial structure of the HSI data.
\subsection{Local linearized patch model construction}\label{sec:HDC_MLP}
Several methods have been proposed to extract the local linear patches from a manifold using K-means clustering \cite{ tan2015grassmann}, \cite{kim2007boosted}, \cite{yang2009video}  and hierarchical agglomerative clustering (HAC) \cite{zhao2007discriminant}, \cite{fan2006locally}. These methods do not consider the linearity property of the extracted local patches during manifold formation. Moreover, the number of extracted local patches are needed to be specified prior to the clustering. Apart from that, the Euclidean distance measure becomes uniform and adversely affects the data representation when dimension is high. In order to overcome these limitations, a concept of maximal linear patch (MLP) \cite{wang2008manifold} with top-down hierarchical divisive clustering (HDC) was proposed in \cite{wang2009manifold}. The principal objective of the MLP lies in two major criteria. 1) linear patch criterion: for every point pair, their geodesic distance must be nearly equal to their euclidean distance, which ensures that the patch lies in the linear subspace; 2) maximal linear patch criterion: the patch size is maximized until the appended data sample violates the linear patch criterion. The nonlinearity degree of this technique is measured by this linear patch criterion or deviation between euclidean distances and geodesic distances \cite{tenenbaum2000global} and \cite{wang2011maximal}. In this work, we choose the hierarchical clustering technique HDC to construct the local linear patches from the nonlinear manifold, due to its ability to construct cluster trees or dendrograms of different degrees. 

We constructed local linear patches ($S_{1}, S_{2}, ... , S_{n}$) out of the filtered training data samples ($X_L$) using HDC-MLP algorithm for each class separately. That means samples from each class is further divided into different patches. Thus, the non-linear manifold is approximated by the union of the local patches, each containing samples from one class only, given by
\begin{equation}
\qquad \mathcal{M} = \bigcup_{k = 1}^n {{S_k}}, \quad \mathrm{and} \quad S_{i} \cap S_{j} = \phi
\end{equation}
where, $L$ is the total number of labeled training data samples and $n$ is the total number of disjoint linear patches.  Lets assume $t_k$ is the number of data samples in the $k^{th}$ patch ($S_k$), so that $\sum\limits_{k = 1}^n {{t_k} = L}$.


The major advantage of generating the local patches include 1) preserving the inherent structure of non-linear manifold, and 2) using these local patches instead of the class samples for construction of the projection matrix in MLSC. Extracting local patches proves to be beneficial for obtaining the projection vectors that can achieve the optimal performance locally.


\subsection{Manifold Local Scaling Cut (MLSC)} \label{sec:MLSC}
In order to extract the spectral-domain information from the generated linear patches, we propose a manifold local scaling cut (MLSC) method. After representing the manifold by the local linear patches, the aim is to construct the optimal projection matrix. MLSC criterion is constructed for the purpose of exploiting both the local linear patch geometry and the global manifold structure of the HSI data. It exploits the geometry by using a Knn graph over extracted local linear patches from the nonlinear manifold.



 
 In \cite{wang2009manifold}, the discriminant function has been calculated by considering the patch centers, which signifies that every patch obeys Gaussian distribution with equal variances. However, real world data usually have non-Gaussian distribution. In order to address this issue, the existing graph-based DR approaches like SC \cite{zhang2009local}, LSC \cite{zhang2015scaling} or SSLSC \cite{zhang2013semisupervised} select the nearest data samples from the same class and from dissimilar classes for constructing the dissimilarity matrix. This has a limitation too; if a sample is surrounded by other class samples in all directions, then proper projection matrix is not learned. However, this can be overcome by considering samples in groups instead of individuals. When samples are grouped, we can use the nearest group of similar and dissimilar classes for constructing the dissimilar matrix. 
 
 MLSC solves the above issue by working on the local linear patches ($S_i$). It optimizes the projection vectors by using discriminant analysis on the samples locally. The formation of local patches facilitates the algorithm to select the closest dissimilar class patches for constructing the dissimilar matrix. Since the patches are locally linear, it guarantees the appropriate learning of projection directions. Moreover, the number of samples in a patch is determined based on the linearity constraints. Thus, the selection of the neighboring patch, instead of neighboring $k$ samples, helps in preserving the data variance and the inherent manifold structure.  These closest patch pairs are determined by computing the inter-patch distances for every patch of one class with every other patch of dissimilar classes only.
 
\begin{figure}[tbp]
	\begin{center}
		\includegraphics [width=0.48\textwidth] {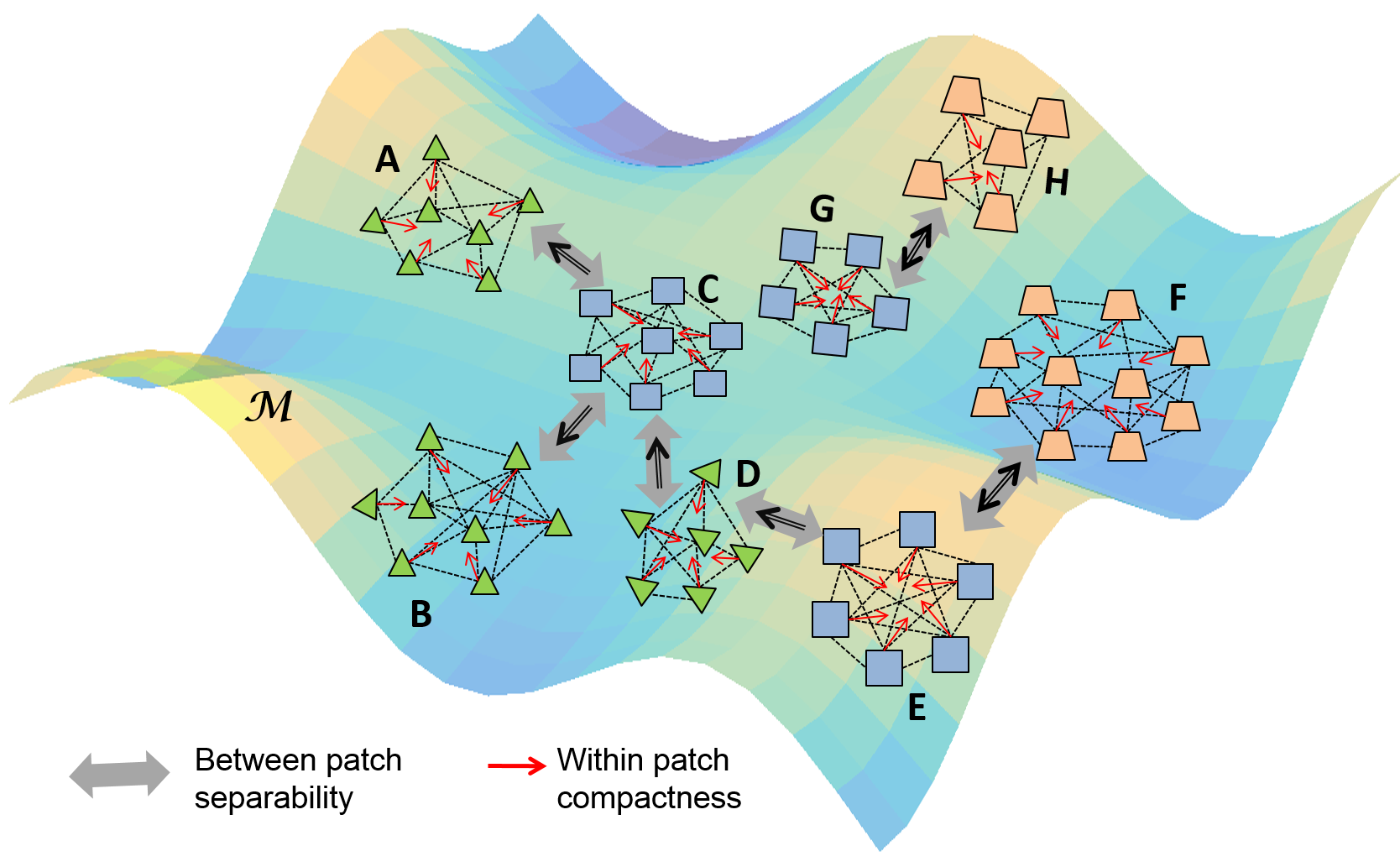}
	\end{center}
    \caption{Illustration of the inter-patch dissimilarity matrix construction for MLSC.}
	\label{fig:mlsc}
\end{figure}

 
 The conceptual illustration of the proposed method is shown in Fig.~\ref{fig:mlsc}. As can be seen in Fig.~\ref{fig:mlsc}, the nonlinear manifold $\mathcal{M}$ is represented by the linear local patches $\{A, B, C, D, E, F, G, H\}$, each of which contains samples of a single class. MLSC uses the data points of the closest patch of a different class to construct the dissimilarity matrix. The between class separability is represented by double headed thick blue arrows. The closest patch pairs based on their inter patch distances are shown by thin black arrows within the thick blue arrows. For instance, $G \Leftrightarrow H$ (\textit{ $\Leftrightarrow$: signifies $H$ is the nearest patch of $G$ and vice-versa}) as distance between $G$ and $H$ is the smallest inter-patch distance for $G$ as well as $H$. Similarly $A \Leftarrow C$ ( \textit{$\Leftarrow$: signifies $C$ is the nearest patch of $A$ but not vice-versa}). Note that $C \Leftarrow D$ as $D$ is the nearest one for $C$. However, $E$ is the nearest one for $D$, therefore, $D \Leftarrow E$. Although the distance between patch $B$ and $D$ are the least, $D$ can't be the nearest patch pair to $B$ because $B$ and $D$ belong to the same class. Hence, $B$ is paired with $C$ and $B \Leftarrow C$. The between patch distances are maximized to enhance their separability (shown in \textit{blue arrow}) and the within patch, data samples are compressed to enhance the within patch compactness (shown in \textit{red arrow}). 

Now, we have the manifold $\mathcal{M} = \{S_{1}, S_{2}, ... , S_{n}\}$ represented by $n$ number of local linear patches. 
Let $S_k $ and $S_{k'} $ be two local patches of different classes which are situated close to each other. Since the nonlinear manifold is locally approximated by the linear patches, it is quite obvious to assume that, the distance between the data samples and patches are locally linear \cite{belkin2001laplacian}. Hence, the distance between two close local patches of the different class is a euclidean distance and it is computed as: 
\begin{equation} \label{clust_dist}
\begin{split}
& {D}^{patch}_{{S_k}-{S_{k'}}} = ||{\mu}^{S_k} - {\mu}^{S_{k'}}||  \,\,\, \text{where}  \\
& {\mu}^{S_k}  = \,\frac{1}{{{t_k}}}\sum\limits_{x_i \in S_k} x_i \,\,  \text{and} \,\, {\mu}^{S_{k'}}  = \,\frac{1}{{{t_{k'}}}}\sum\limits_{x_i \in S_{k'}} x_i
\end{split}
\end{equation}


\begin{equation} \label{eq:spec_bet_withn_dis_mat}
\begin{split}
B^{Spec}_k\, = \,\sum\limits_{x_i\, \in \,\,{S_k}} {\sum\limits_{x_j\, \in \,\,{S_{k'}}} {\frac{1}{{{t_k}{t_{k'}}}}({x_i}\, - \,{x_j}){{({x_i}\, - \,{x_j})}^T}} }  \\
W^{Spec}_k\, = \,\sum\limits_{x_i\, \in \,\,{S_k}} {\sum\limits_{x_j\, \in \,\,{S_k}} {\frac{1}{{{t_k}{t_k}}}({x_i}\, - \,{x_j}){{({x_i}\, - \,{x_j})}^T}} } 
\end{split}
\end{equation}
where $t_k$ and $t_{k'}$ denotes the number of samples in local patch $S_k$ and $S_{k'}$ respectively. $B^{Spec}_k$ measures the dissimilarity matrix between samples in local $k_{th}$ patch $S_k$ and its nearest neighbor $S_{k'}$. Similarly, $W^{Spec}_k$ measures the total dissimilarity within the samples of the patch $S_k$. Here, we are only using the samples of closest local linear patch pairs of different classes instead of all samples of the classes for computing the dissimilarity matrix. For example, in Fig.~\ref{fig:mlsc}, local linear patches $\{A, B, D\}$, $\{C,E,G\}$ and $\{F,H\}$ are having different class labels. We can observe that patch $B$ and $D$ are more close to each other, than $\{B,C\}$ or $\{C,D\}$ or $\{D,E\}$. However, $S_{k'}=C$ for $S_k=B$ as $B$ and $D$ belong to the same class whereas patch $C$ belongs to a different class. Similarly, we can say $S_{k'}=E$ for $S_k=D$. Then, these nearest patches of the varied class are used for constructing the dissimilarity matrix. These dissimilarity matrices are used in the optimization process to obtain the optimal projection matrix for dimension reduction. 
Using the definition of $B^{Spec}_k$ and $W^{Spec}_k$ the objective function of MLSC is defined as 
\begin{equation} \label{eq:mlsc_optmiz_prblm}
\begin{split}
MLSC(V)\, & = \,\frac{{\left| {\sum\limits_{k = 1}^n {{V^T}\,{B^{Spec}_k}\,V} } \right|}}{{\left| {\sum\limits_{k = 1}^n {({V^T}\,{W^{Spec}_k}\,V\, + \,{V^T}\,{B^{Spec}_k}\,V)} } \right|}} \\
& = \,\frac{{\left| {{V^T}\,{B^{Spec}}\,V} \right|}}{{\left| {{V^T}\,({W^{Spec}} + {B^{Spec}})\,V} \right|}} 
 = \,\frac{{\left| {{V^T}\,{B^{Spec}}\,V} \right|}}{{\left| {{V^T}\,{T^{Spec}}\,V} \right|}} \\
\makebox{and} \\
& \sum\limits_{k = 1}^n {B^{Spec}_k} = B^{Spec} \, ; \, 
\sum\limits_{k = 1}^n {W^{Spec}_k} = W^{Spec} \,
\end{split}
\end{equation}
where $B^{Spec}$ is between-patch, $W^{Spec}$ is within-patch and $T^{Spec} = B^{spec} + W^{spec}$ is the total dissimilarity matrix in spectral domain of all local patches. These spectral dissimilarity matrices are used to construct the optimal projection matrix $V$ by simultaneously maximizing the between-patch dissimilarity matrix and minimizing the within-patch dissimilarity matrix.

\subsection{Spectral Regularized Manifold LSC (RMLSC)} \label{sec:RMLSC}
The major limitation of spectral MLSC method is that, it suffers from singularity issue caused due to small size samples. The above limitations are addressed by appending a regularized term in the spectral MLSC. The regularizer also increases the patch discrimination by enhancing the inter-patch variability. The newly derived criterion is termed as regularized manifold LSC (RMLSC). The RMLSC performs the spectral-domain manifold local scaling cut operation with a penalty term. Inspired by the existing literatures on spectral domain DR methods in \cite{zhou2015dimension} and \cite{zhang2015scaling}, we propose a new objective function for the RMLSC criteria. The objective function is defined as

\begin{equation} \label{eq:regularizer_MLSC}
\begin{split}
RB^{spec} = & tr(V^T [(1-\alpha)B^{spec} + \alpha XX^T] V); \\
RW^{spec} = & tr(V^T [(1-\alpha)W^{spec} + \alpha (\mbox{\textit{Diag}}(diag(W^{spec})))] V) \\
RT^{spec} = & RB^{spec} + RW^{spec} \\
    = & tr(V^T[(1-\alpha)(W^{spec} + B^{spec}) + \alpha (R_w+R_b)]V) \\
    \end{split}
\end{equation}
\begin{equation} \label{eq:RMLSC_optz_func}
\begin{split}
RMLSC(V) &= \mathop {\max }\limits_{V\, \in \,{R^{D \times d}}} \, \frac{RB^{spec}}{RT^{spec}}\\        
 = \mathop {\max }\limits_{V\in {R^{D \times d}}} & \frac{tr(V^T [(1-\alpha)B^{spec} + \alpha R_b] V)}{tr(V^T[(1-\alpha)(T^{spec}) + \alpha (R_{w}+{R_b})] V)}
\end{split}
\end{equation}
where $R_{w} = \mbox{\textit{Diag}}(diag(W^{spec})))$ and $R_{b} = XX^T$ are the regularizers of the within-patch dissimilarity $W^{spec}$ and between-patch dissimilarity $B^{spec}$ respectively. $\alpha \in [0,1]$ is the regularization parameter, $tr(\cdot)$ is the trace of a matrix, $diag(\cdot)$ represents the diagonal elements of a matrix, and $\mbox{\textit{Diag}}(\cdot)$ converts the vector into a diagonal matrix. The numerator ($RB^{spec}$) of the objective function corresponds to the between-patch dissimilarity with the regularization term $R_b$ and the denominator ($RT^{spec} = RB^{spec} + RW^{spec} $) represents the combination of within-patch ($RW^{spec}$) and between-patch ($RB^{spec}$) dissimilarity matrix with their corresponding regularizer $R_w$ and $R_b$.

The major modification in this RMLSC is the regularization terms $R_b$ and $R_w$. 
The regularization term $R_b$ in the numerator well preserve the data diversity by maximizing the data variance \cite{zhou2015dimension}, \cite{weinberger2006introduction}. It is proven that, the classification performance is greatly improved by well preserved data diversity \cite{gao2013stable}. The regularizer $R_w$ used in the within-patch dissimilarity matrix is a diagonal element, and it improves the stability of the solution. Due to limited training samples in HSI \cite{liao2013semisupervised}, the eigenvalues decay very rapidly to zero \cite{jiang2008eigenfeature}, \cite{zhou2015dimension}. These Small or zero eigenvalues attain instability and lose discriminative information by placing null spaces in the basis. This diagonal regularizer reduces the decay of the eigenvalues by acting against the bias estimation of small eigenvalues of the limited training data \cite{friedman1989regularized}. 
Hence, it provides better stability. However, in the denominator, the total dissimilarity combines the both $RB^{spec}$ and $RW^{spec}$. Hence, the denominator provides both the data diversity and stability to the solution.

When $\alpha = 0$, the RMLSC becomes the MLSC. The RMLSC uses the labeled samples to determine the discriminative projection direction by considering the original distribution and modality. The regularization terms are added to the between-patch and within-patch dissimilarity matrix to incorporate the data diversity and stability in the local manifold structure of the neighborhood samples.


\subsection{Spatial Neighboring Pixel MLSC (NPMLSC)}
In spatial-domain, the neighboring pixels share similar land cover properties and usually belong to the same class. Hence, this spatial information of the MLSC patches can be useful in determining the projection matrix to improve the classification performance. In the proposed neighboring pixel manifold local scaling cut (NPMLSC) method, we construct a dissimilarity matrix using the spatial neighborhood pixel information of the MLSC patches. It preserves the original spatial neighborhood pixel correlation in the projected NPMLSC embedding space.


As explained in section~\ref{sec:MLSC}, we first determine the locally situated nearest linear patch of $S_k$, i.e. $S_{k'}$. Here, both $S_k$ and $S_{k'}$ are of two different classes. Next, we determine the spatial neighborhood of all the pixels present in $S_{k'}$. Let $x_j$ be a pixel in patch $S_{k'}$ ($x_j \in S_{k'}$). Let's denote the $p$ surrounding pixels in the spatial neighborhood of $x_j$ by $P_j = \{x_{j1}, x_{j2}, ...,x_{jp}\}$. The total number of spatial neighborhood elements in $S_{k'}$ is denoted as $\bar{S_{k'}}$ and $\bar{S_{k'}} = \{P_1,P_2,...\} = \{x_{11}, x_{12}, ...,x_{1p},x_{21}, x_{22}, ...,x_{2p}, ..., x_{j1}, x_{j2}, ...,x_{jp},...\}$. 
Then, using these spatial neighborhood elements of the nearest patch $\bar{S_k}$ and $\bar{S_{k'}}$, we compute both the between-patch dissimilarity matrix ($B^{spa}$) and within-patch dissimilarity matrix ($W^{spa}$). 
\begin{align} \label{eq:spat_bet_withn_disim_mat}
\begin{split}
B^{spa} = \sum_{k =1}^{n}\sum_{x_i \in \bar{S_k}}\sum _{x_j \in \bar{S_{k'}}} \eta_{ij} ({x_i} - x_{j})({x_i} - x_{j})^T \\ 
W^{spa} = \sum_{k =1}^{n}\sum_{x_i \in \bar{S_k}} \sum _{x_j \in \bar{S_{k}}, x_i\ne x_j} \eta_{ij} ({x_i} - x_{j})({x_i} - x_{j})^T
\end{split}
\end{align}
where $\eta_{ij} = \frac{d_{ij}}{\sum_{ (i,j)} d_{ij}}$ and $d_{ij} = \exp(- \gamma ||(x_i - x_{j})||^2)$
is a weight function, which controls the effect of contribution of the points based on their spectral distance measure. The NPMLSC seeks the linear projection matrix to maximize the spatial neighborhood class discrimination using $B^{spa}$ and $W^{spa}$.


\subsection{ Spatial Spectral Regularized MLSC (SSRMLSC)}\label{ssec:SSRMLSC}
Due to the heteroscedastic distribution of HSI data, the spectrally close samples may belong to different classes. Hence, only spectral distance measure is inadequate for determining the optimal projection matrix. The spectral RMLSC method exploits the local intrinsic manifold of the data in spectral domain. 
On the other hand, NPMLSC uses the spatial information to retain the local pixel neighborhood structure of the linear patch without using the labeled spectral information. However, NPMLSC fails to connect two pixels with spatially higher pixel distance in a homogeneous region or in a linearly constructed patch. In such a case, the labeled spectral information plays a vital role to establish a connection, which improves the discrimination criteria.

Therefore, labeled spectral information and spatial information complement each other in terms of information content and thereby improves the HSI classification performance. Here, We incorporate the information from the spatial domain with the spectral domain and propose a spatial-spectral RMLSC (SSRMLSC) method. By merging the spectral RMLSC and spatial NPMLSC method, we construct spatial-spectral between-patch dissimilarity matrix $B^{SS}$ and within-patch dissimilarity matrix $W^{SS}$ as
\begin{equation}\label{eq:spat-spec_bet_withn_disim_mat}
\begin{split}
W^{SS} = \beta (RW^{spec}) &+ (1 - \beta)W^{spa} \\
B^{SS} = \beta (RB^{spec}) &+ (1 - \beta)B^{spa} \\
T^{SS} = W^{SS} &+ B^{SS}
\end{split}
\end{equation}
where $\beta \in [0,1]$ is used to control the contribution of the spectral and spatial information.  The optimal projection matrix $\dddot{V}$ 
is obtained by solving by the generalized eigenvalue problem. The obtained projection matrix project the original data to a lower dimensional space spanned by $\dddot{V}$ to get the new feature vectors. 

\subsection{Semi-supervised Spatial Spectral Regularized MLSC (S3RMLSC)}\label{ssec:S3RMLSC}
SSRMLSC seeks the optimal projection matrix for dimension reduction by purely using the labeled training data in both spectral and spatial domains. However, large amount of unlabeled data are available in practical scenarios. Semi-supervised algorithms take the advantage of the abundance of the unlabeled data to improve classification performance. The proposed SSRMLSC algorithm can be extended to semi-supervised spatial spectral regularized MLSC (S3RMLSC) method by adding a penalty term derived from the unlabeled data. This exploits the underlying geometrical structure of the unlabeled data during construction of projection vectors. 

For this semi-supervised method, we construct a weighted undirected graph $G \in (X,E)$ from the total training dataset of $X = (X_L, X_U)$. Each observation in the dataset is considered as a node and they are connected by a set of edges $E$ with some associated weights. These weighted edges are represented by an adjacency matrix.  
The adjacency matrix $\mathcal{A}$ of the graph $G$ is determined by computing the Knn of each vertex (data sample). The diagonal elements of the computed adjacency matrix are zero as the distance measure is zero for the same vertex. Then the graph laplacian $\mathcal{L}$ is determined by
\begin{equation} \label{eq:laplacian_mat}
\mathcal{L} = \mathcal{D} - \mathcal{A}
\end{equation}
where $\mathcal{D}$ is the diagonal degree of the adjacency matrix $\mathcal{A}$. This degree estimates the density around the data samples. The $i$th diagonal entry of $\mathcal{D}$ is calculated by $\mathcal{D}_{ii} = \sum_{j=1}^{L+U}a_{ij}$, where $a_{ij}$ is an element of adjacency matrix $\mathcal{A}$. Here both adjacency and laplacian matrices are symmetric in nature.  $V$ is the optimal projection matrix for DR such that $z_i = V^{T}x_i \in R^{d}$. If we consider two close points $x_i$ and $x_j$ in a manifold, then their projection vectors $z_i$ and $z_j$ are expected to be as close as possible on the reduced dimensional hyperplane.
Hence, the projection matrix can be obtained by solving the optimization with respect to $V$ in the equation, 
\begin{equation}
min{\sum_{i,j}a_{ij}{||z_i - z_j||}_2^2}
\end{equation} 
Motivated by \cite{zhang2013semisupervised} and \cite{belkin2006manifold}, we obtain a proper projection matrix $V^*$ by formulating the regularization term as $V^TX{\mathcal{L}}X^TV$. To make this paper self-contained, we derive the regularization term as 
\begin{equation} \label{eq:semSup_proof}
\begin{split}
& \frac{1}{2}{\sum_{i=1,j=1}^{L+U}a_{ij}{||z_i - z_j||}_2^2}
 =\frac{1}{2}\sum\limits_h {\sum\limits_{i=1,j=1}^{L+U} {{a_{ij}}{{(v_h^T{x_i} - v_h^T{x_j})}^2}} }\\
& = \sum\limits_h {\left( {\sum\limits_{i=1,j=1}^{L+U} {v_h^T{x_i}{a_{ij}}x_i^T{v_h} - \sum\limits_{i=1,j=1}^{L+U} {v_h^T{x_i}{a_{ij}}x_j^T{v_h}} } } \right)}   \\
& =\sum\limits_h {v_h^T\left( {\sum\limits_{i=1}^{L+U} {{x_i}{d_{ii}}x_i^T - \sum\limits_{i=1,j=1}^{L+U} {{x_i}{a_{ij}}x_j^T} } } \right){v_h}} \\
& = \sum\limits_h {v_h^TX(\mathcal{D} - {\mathcal{A}})X^T{v_h}} 
 = tr\left( {{V^T}X(\mathcal{D} - {\mathcal{A}}){X^T}V} \right)\\
&=tr\left( {{V^T}X{\mathcal{L}}{X^T}V} \right)
\end{split}
\end{equation}
We obtain the updated objective function for the S3RMLSC by adding the penalty term with SSRMLSC. Motivated by \cite{zhang2015scaling}, we formulate this optimization problem as a trace-ratio problem:
\begin{equation} \label{eq:sssRMLSC_trace_ratio}
V^* \,  =  {arg}\, \mathop {\max }\limits_{V\, \in \,{R^{D \times d}}} \,{\frac{{tr\left( {{V^T}{B^{SS}}V} \right)}}{{tr\left( {{V^T}\left( {{T^{SS}} + \,\gamma X{\mathcal{L}}{X^T}} \right)V} \right)}}}
\end{equation} 
where $\gamma$ is called a pooling parameter to balance the contribution of the regularization term and it ranges between $[0,1]$. 

Here, we use the trace-difference problem proposed in \cite{wang2007trace}  and \cite{jia2009trace} to solve the trace-ratio problem. 
Hence, we formulate the trace-ratio problem as a trace-difference problem:
\begin{equation} \label{eq:sssRMLSC_trace_diff}
\begin{split}
V^* \,  = &  {arg}\, \mathop {\max }\limits_{V\, \in \,{R^{D \times d}}} \,  \\
& {tr({{{V^T}\,({B^{SS}} -\lambda \times{({{T^{SS}} + \,\gamma X{\mathcal{L}}{X^T}})})\,V}})}
\end{split}
\end{equation} 
This trace-difference problem has been solved by a technique called \textit{decomposed Newton method} (DNM) \cite{wang2007trace} to achieve the global optimum of the trace-ratio problem. In a general way, the trace-difference function depends on the largest $d$ eigenvalues. Initially, it determines the eigenvalue set by function decomposition and use their Taylor series expansion to approximate the function value. Then, it finds the eigenvalue $\lambda$ by solving this approximated function in an iterative manner.\\ 
Since these dissimilarity matrices are scaled by the size of the patches; therefore, this graph cut criterion is called scaling cut criterion. As this scaling cut criterion is applied between two nearest patches, it is termed as \textit{localized scaling cut} criterion. 

The obtained optimal projection matrix $V$ in S3RMLSC, extracts reliable boundary of the linear patches in the manifold space. We use the obtained projection matrix $V$ over the labeled training set and testing dataset to project it onto the new reduced dimension. Later, support vector machine (SVM) classifier is employed on the projected test dataset to predict the labels for evaluating its accuracy. 

\section{Experimental Results and Analysis} \label{sec:expt}


\subsection{Dataset description}
This paper adopts two benchmark real world HSI datasets, i.e., Indian pine \cite{PURR1947} and Botswana to conduct our experiments.

i) Indian pine dataset: It was captured by airborne visible/imaging spectrometer over Northwest Indiana's Indian pine test site. The size of the dataset is $145 \times 145$ in the spatial direction and it contains $200$ bands in spectral direction with $16$ ground truth classes. 

ii) Botswana dataset: It is a spaceborne dataset collected over Okavango Delta, Botswana. This was obtained by Hyperion sensor of NASA Earth observing-1 satellite. This image has $1476 \times 256$ spatial pixels and $145$ spectral bands with $14$ ground truth classes.  

\begin{table}[!b]
\centering
\caption{Overall accuracy (\%) with varying $w$.}
\label{tab:spatial_window_OA}
\resizebox{0.49\textwidth}{!}{%
\begin{tabular}{c|cccccccc}
\hline
w    &  $3\times3$ & $5\times5$ & $7\times7$ & $9\times9$ & $11\times11$ & $13\times13$ & $15\times15$ \\ \hline
Botswana &        \textbf{97.48}      &     96.36       &     97.28       &     96.80       &      95.45        &  96.96            &      96.99        \\ 
Indian Pines &        \textbf{93.84}       &    93.16        &    93.64        &    93.06        &    93.45          &   93.09           &  92.77            \\ \hline
\end{tabular}%
}
\end{table}

\begin{table*}[b]
\centering
\caption{The effect of the number of unlabeled data in semi-supervised training (OA in \%).}
\label{tab:ULdata_OA}
\begin{tabular}{c|llllllllllllll}
\hline
UL Data points   &  $200$ & $400$ & $600$ & $800$ & $1000$ & $1200$ & $1400$ & $1600$ & $1800$ & $2000$ & $2200$ & $2400$ & $2500$\\ \hline
Botswana &        97.30      &     96.78       &    97.15       &     96.32       &      96.97        &  96.61            &      97.23  &  96.54 &  96.79   &  \textbf{97.91}  &   96.98   &  97.18  &   97.10      \\ 
Indian Pines &        93.44       &    93.72        &    93.02        &    92.76        &    93.02          &   93.07           &  93.41  &  93.50  &  92.78  & \textbf{93.98}  &  93.93  & 93.33  &   92.74          \\ \hline
\end{tabular}%
\end{table*}


\subsection{Experimental settings}
We provide a comparative analysis of the performance of the proposed approaches with the state-of-the-art techniques in this section. 
SVM without DR is reported which acts as the baseline. The supervised methods, such as SC, LSC, NWFE, RLDE, and MLSC use the labeled pixels to compute the projection matrix. For supervised spatial-spectral methods, such as SSRLDE, SSRMLSC, and S3RMLSC, we used the labeled pixels and their corresponding spatial pixels to compute the projection matrix. However, for the semi-supervised methods like SELDLPP, SELDNPE \cite{liao2013semisupervised}, SSLSC, and S3RMLSC, the complete training set (labeled + unlabeled pixels) are used. For a fair comparison and to understand the effect of guided filtering, we have considered the proposed technique and its variants in both with filtering (HGF) and without filtering (No HGF) conditions. The semi-supervised SSRLDE technique used in the comparison is considered with filtering (weighted mean filtering (WMF)) condition only. 
The evaluation of S3RMLSC is carried out by determining the class accuracy (CA), overall classification accuracy (OA), class average accuracy (AA), and kappa coefficient ($k$) \cite{congalton2008assessing} of the SVM with the linear kernel as the classifier on the projected data. All obtained results are the average of five independent iterations. The parameters of the comparative approaches are selected based on their relevant literature. For both the datasets, we randomly selected $10\%$ labeled pixels from each class and $2000$ unlabeled pixels to construct the training set for the semi-supervised case and the remaining pixels as the testing set. 

\subsection{Parameter selection}
S3RMLSC uses several basic parameters such as - regularization parameter $\epsilon$ in HGF, spectral penalty parameter $\alpha$, spatial-spectral contribution pooling parameter $\beta$, spatial neighborhood pixels $p$, unlabeled data samples $u$, Knn graph parameter for Laplacian matrix $k$, and semi-supervised regularization parameter $\gamma$. The experimental value of the regularization parameter of HGF is set to $\epsilon = 0.001$. This parameter is selected based on its parental literature \cite{pan2017hierarchical}. To obtain the suitable value of $\alpha$ and $\beta$, we performed a grid search while varying both parameters in the range [0,1]. Empirically we found the values $(\alpha= 0.4)$ and $( \beta = 0.3)$ to be the best fit for the problem at hand. 

In the experiment, $p$ is related to a spatial neighborhood square window $w$. Having a $w = 3 \times 3$ window results with $p = 9$. 
We varied $w$ from $3 \times 3$ to $15 \times 15$, and obtained maximum OA at window size $3 \times 3$ as shown in Table~\ref{tab:spatial_window_OA}. The results are achieved with $10\%$ labeled training samples per class, dimensions $30$, $\alpha = 0.4$, and $\beta = 0.3$. Larger window size increases the probability of interference from the pixel of other classes. Hence, we select the window $w$ of size $3 \times 3$ to reduce the interference and it is used in subsequent experiments.

To show the impact of the number of unlabeled samples in semi-supervised training, we experimented with different unlabeled data size in the range $\{200,400, \cdots 2500\}$ as shown in Table~\ref{tab:ULdata_OA}. These experiments are conducted by selecting $10\%$ labeled samples per class as training data and the rest data as the test set. The unlabeled training data samples are randomly selected from the test set. As per the Table~\ref{tab:ULdata_OA}, we found that the proposed approach performs well while using $2000$ unlabeled samples in training for both the datasets. Hence we selected $2000$ number of unlabeled samples for the rest of the experiments.


\begin{figure}[h!]
	\centering
		\includegraphics[width=.55\linewidth]{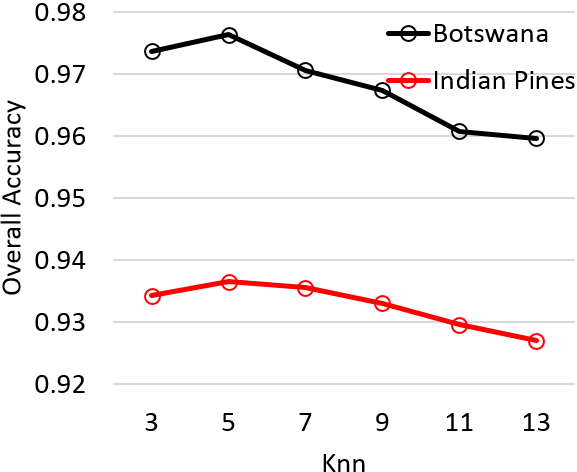}
	\caption{Effect of Knn parameter on OA of S3RMLSC.}
	\label{fig:OA_knn_IndP_Bots}
\end{figure}

Empirically we set the regularization parameter $\gamma$ to $0.5$. Fig.~\ref{fig:OA_knn_IndP_Bots} shows the variations of OA with respect to the various $k$ value of Knn in the proposed semi-supervised approach. From the Fig.~\ref{fig:OA_knn_IndP_Bots}; we found that the proposed method performs well for $k = 5$. Hence, we consider $k =5$ for the rest of our experiments. 

\begin{table*}[htbp]
\centering
\caption{Comparison of the best classification accuracies (in \%) with corresponding dimensions (in bracket) and computation time (in sec) of Indian Pines dataset. 10\% of labeled samples per class and 2000 unlabeled samples are used in this experiment. 
}
\label{tab:res_IndP_data}
\resizebox{0.97\textwidth}{!}{
\begin{tabular}{c|cc|cccccc|c|c|c|c|c|ccc|c|c}
\hline
\multirow{2}{*}{Class} & \multirow{2}{*}{\begin{tabular}[c]{@{}c@{}}Train \\ size\end{tabular}} & \multirow{2}{*}{\begin{tabular}[c]{@{}c@{}}Test\\ size\end{tabular}} & \multirow{2}{*}{\begin{tabular}[c]{@{}c@{}}RAW \\ (200)\end{tabular}} & \multirow{2}{*}{\begin{tabular}[c]{@{}c@{}}SC \\  (48)\end{tabular}} & \multirow{2}{*}{\begin{tabular}[c]{@{}c@{}}LSC \\ (48)\end{tabular}} & \multirow{2}{*}{\begin{tabular}[c]{@{}c@{}}LFDA\\ (50)\end{tabular}} & \multirow{2}{*}{\begin{tabular}[c]{@{}c@{}}NWFE\\ (36)\end{tabular}} & \multirow{2}{*}{\begin{tabular}[c]{@{}c@{}}RLDE\\ (38)\end{tabular}} & \multicolumn{2}{c|}{MLSC}                                       & SSRLDE                                             & \multicolumn{2}{c|}{SSRMLSC}                                    & \multirow{2}{*}{\begin{tabular}[|c]{@{}c@{}}SELDLPP\\ (44)\end{tabular}} & \multirow{2}{*}{\begin{tabular}[c]{@{}c@{}}SELDNPE\\ (42)\end{tabular}} & \multirow{2}{*}{\begin{tabular}[c]{@{}c@{}}SSLSC\\ (16)\end{tabular}} & \multicolumn{2}{c}{S3RMLSC}                                                        \\ \cline{10-11} \cline{13-14} \cline{18-19}
                       &                                                                        &                                                                      &                                                                       &                                                                      &                                                                      &                                                                      &                                                                      &                                                                      & \begin{tabular}[c]{@{}c@{}}HGF\\ (50)\end{tabular} & \begin{tabular}[c|]{@{}c@{}}No HGF\\ (46)\end{tabular} & \begin{tabular}[|c]{@{}c@{}}WMF\\ (42)\end{tabular} & \begin{tabular}[c]{@{}c@{}}HGF\\ (48)\end{tabular} & \begin{tabular}[c]{@{}c@{}}No HGF\\ (50)\end{tabular} &                                                                         &                                                                         &                                                                       & \begin{tabular}[c]{@{}c@{}}HGF\\ (48)\end{tabular} & \begin{tabular}[c]{@{}c@{}}No HGF\\ (46)\end{tabular} \\ \hline 
1 & 10 & 36 & 53.89 & 61.11 & 61.24 & 55.00 & 65.56 & 56.67 & 95.16 & 66.58 & 30.65 & 98.49 & 88.17 & 27.32 & 20.23 & 45.56 & 97.00 & 75.31 \\
2 & 143 & 1285 & 73.45 & 70.57 & 74.69 & 65.85 & 74.91 & 66.85 & 92.07 & 69.53 & 82.79 & 93.02 & 67.83 & 87.81 & 68.67 & 54.10 & 94.48 & 69.53 \\
3 & 83 & 747 & 46.79 & 52.82 & 56.98 & 48.81 & 55.37 & 50.63 & 94.91 & 73.82 & 75.94 & 97.40 & 57.66 & 75.42 & 66.34 & 44.87 & 97.42 & 65.82 \\
4 & 24 & 213 & 58.12 & 52.77 & 52.88 & 42.35 & 63.66 & 53.33 & 95.75 & 57.09 & 87.21 & 96.97 & 73.36 & 69.09 & 69.21 & 43.85 & 97.53 & 77.09 \\
5 & 49 & 434 & 87.70 & 84.61 & 84.73 & 88.76 & 87.56 & 88.29 & 93.84 & 84.33 & 90.00 & 95.22 & 82.48 & 87.21 & 71.49 & 87.24 & 95.77 & 84.53 \\
6 & 73 & 657 & 90.31 & 95.53 & 95.65 & 94.89 & 95.56 & 94.22 & 99.25 & 93.18 & 82.92 & 98.35 & 94.67 & 91.8 & 93.45 & 93.94 & 98.54 & 93.18 \\
7 & 10 & 18 & 85.56 & 84.44 & 84.57 & 92.22 & 91.11 & 85.56 & 94.56 & 84.44 & 73.29 & 95.16 & 86.67 & 34.54 & 30.16 & 76.67 & 94.22 & 84.44 \\
8 & 48 & 430 & 85.81 & 93.91 & 94.03 & 97.12 & 98.60 & 95.95 & 99.18 & 95.53 & 97.49 & 99.76 & 92.84 & 98.32 & 99.55 & 98.33 & 98.39 & 95.53 \\
9 & 10 & 10 & 80.00 & 86.00 & 86.13 & 86.00 & 90.00 & 70.00 & 99.89 & 86.00 & 28.90 & 100.00 & 82.00 & 25.12 & 22.34 & 70.00 & 100.00 & 86.32 \\
10 & 98 & 874 & 66.89 & 67.39 & 76.49 & 44.55 & 66.16 & 63.43 & 89.93 & 76.31 & 89.20 & 90.35 & 76.16 & 86.31 & 84.57 & 59.24 & 91.63 & 68.31 \\
11 & 246 & 2209 & 81.28 & 79.67 & 79.79 & 88.20 & 81.96 & 71.30 & 97.09 & 80.53 & 99.86 & 98.59 & 77.71 & 96.18 & 98.70 & 81.11 & 98.26 & 80.53 \\
12 & 60 & 533 & 64.02 & 55.38 & 55.51 & 52.57 & 67.88 & 62.96 & 96.29 & 64.71 & 95.90 & 97.88 & 79.38 & 84.35 & 88.92 & 36.21 & 97.78 & 69.71 \\
13 & 21 & 184 & 95.11 & 93.37 & 93.50 & 97.28 & 90.76 & 95.98 & 96.67 & 92.93 & 91.43 & 94.14 & 93.80 & 88.78 & 72.71 & 93.80 & 93.84 & 92.93 \\
14 & 127 & 1138 & 73.92 & 93.55 & 93.68 & 96.82 & 95.17 & 90.72 & 98.99 & 94.59 & 99.23 & 99.44 & 92.02 & 94.75 & 98.82 & 95.10 & 99.51 & 94.59 \\
15 & 39 & 347 & 65.59 & 60.40 & 58.90 & 58.90 & 62.65 & 60.35 & 96.65 & 72.54 & 97.68 & 97.77 & 57.93 & 68.78 & 89.17 & 58.33 & 97.73 & 69.95 \\
16 & 10 & 83 & 52.41 & 82.89 & 72.05 & 72.05 & 76.39 & 76.63 & 93.80 & 83.37 & 59.76 & 94.35 & 78.55 & 60.4 & 50.54 & 77.83 & 95.39 & 83.37 \\ \hline 
AA & \textbf{1051} & \textbf{9198} & 70.43 & 75.90 & 77.59 & 73.84 & 78.96 & 73.93 & \textbf{95.88} & \textbf{79.72} & 80.14 & \textbf{96.62} & \textbf{80.17} & 73.51 & 70.30 & 69.76 & \textbf{96.72} & \textbf{80.70} \\
OA &  &  & 71.51 & 76.29 & 78.88 & 75.40 & 79.08 & 73.44 & \textbf{95.92} & \textbf{79.95} & 78.53 & \textbf{96.75} & \textbf{80.56} & 75.62 & 73.28 & 71.80 & \textbf{96.86} & \textbf{80.94} \\
$\kappa$ &  &  & 69.34 & 72.77 & 74.32 & 71.35 & 75.94 & 69.63 & \textbf{95.13} & \textbf{75.91} & 75.92 & \textbf{96.19} & \textbf{76.89} & 73.72 & 68.86 & 65.98 & \textbf{96.27} & \textbf{78.91} \\
Time &  &  &  & 931.70 & 20.20 & 0.34 & 216.13 & 126.77 & \textbf{2.06} & \textbf{1.99} & 130.32 & \textbf{5.92} & \textbf{5.78} & 1.71 & 2.49 & 25.06 & \textbf{6.26} & \textbf{6.13} \\ \hline
\end{tabular}}
\end{table*}

\begin{table*}[htbp]
\centering
\caption{Comparison of the best classification accuracies (in \%) with corresponding dimensions (in bracket) and computation time (in sec) of Botswana dataset. 10\% of labeled samples per class and 2000 unlabeled samples are used in this experiment. 
}
\label{tab:res_Bots_data} 
\resizebox{0.97\textwidth}{!}{
\begin{tabular}{c|cc|cccccc|c|c|c|c|c|ccc|c|c}
\hline
\multirow{2}{*}{Class} & \multirow{2}{*}{\begin{tabular}[c]{@{}c@{}}Train\\ Size\end{tabular}} & \multirow{2}{*}{\begin{tabular}[c]{@{}c@{}}Test\\ Size\end{tabular}} & \multirow{2}{*}{\begin{tabular}[c]{@{}c@{}}RAW\\ (145)\end{tabular}} & \multirow{2}{*}{\begin{tabular}[c]{@{}c@{}}SC\\ (44)\end{tabular}} & \multirow{2}{*}{\begin{tabular}[c]{@{}c@{}}LSC\\ (44)\end{tabular}} & \multirow{2}{*}{\begin{tabular}[c]{@{}c@{}}LFDA\\ (50)\end{tabular}} & \multirow{2}{*}{\begin{tabular}[c]{@{}c@{}}NWFE\\ (50)\end{tabular}} & \multirow{2}{*}{\begin{tabular}[c]{@{}c@{}}RLDE\\ (8)\end{tabular}} & \multicolumn{2}{c|}{MLSC} & SSRLDE & \multicolumn{2}{c|}{SSRMLSC} & \multirow{2}{*}{\begin{tabular}[c]{@{}c@{}}SELDLPP\\ (46)\end{tabular}} & \multirow{2}{*}{\begin{tabular}[c]{@{}c@{}}SELDNPE\\ (38)\end{tabular}} & \multirow{2}{*}{\begin{tabular}[c]{@{}c@{}}SSLSC\\ (48)\end{tabular}} & \multicolumn{2}{c}{S3RMLSC} \\ \cline{10-11} \cline{13-14} \cline{18-19}
 &  &  &  &  &  &  &  &  & \begin{tabular}[c]{@{}c@{}}HGF\\ (50)\end{tabular} & \begin{tabular}[c|]{@{}c@{}}No HGF\\ (36)\end{tabular} & \begin{tabular}[|c]{@{}c@{}}WMF\\ (42)\end{tabular} & \begin{tabular}[c]{@{}c@{}}HGF\\ (48)\end{tabular} & \begin{tabular}[c]{@{}c@{}}No HGF\\ (48)\end{tabular} &  &  &  & \begin{tabular}[c]{@{}c@{}}HGF\\ (40)\end{tabular} & \begin{tabular}[c]{@{}c@{}}No HGF\\ (48)\end{tabular} \\ \hline 
1 & 27 & 243 & 85.17 & 96.81 & 97.94 & 99.42 & 99.47 & 99.18 & 99.67 & 99.00 & 97.86 & 100.00 & 98.68 & 97.20 & 97.28 & 99.26 & 99.92 & 98.68 \\
2 & 11 & 90 & 90.78 & 88.65 & 89.78 & 94.00 & 90.89 & 92.89 & 93.01 & 91.57 & 90.00 & 95.13 & 93.78 & 87.11 & 74.22 & 92.44 & 96.31 & 94.00 \\
3 & 26 & 225 & 89.67 & 95.23 & 96.36 & 97.69 & 98.93 & 99.02 & 100.00 & 97.33 & 95.58 & 100.00 & 96.53 & 97.78 & 99.29 & 99.11 & 100.00 & 96.18 \\
4 & 22 & 193 & 89.64 & 94.52 & 95.65 & 94.30 & 97.10 & 96.06 & 98.42 & 91.30 & 90.21 & 98.45 & 92.85 & 98.65 & 86.32 & 93.89 & 98.26 & 92.85 \\
5 & 27 & 242 & 92.15 & 85.73 & 86.86 & 79.17 & 89.17 & 82.89 & 90.94 & 87.63 & 94.17 & 96.48 & 93.69 & 79.92 & 79.17 & 86.78 & 96.25 & 94.02 \\
6 & 27 & 242 & 83.47 & 79.53 & 80.66 & 74.38 & 82.73 & 85.87 & 95.99 & 89.69 & 92.00 & 96.32 & 92.65 & 77.44 & 65.62 & 84.05 & 96.73 & 92.31 \\
7 & 26 & 233 & 87.42 & 95.97 & 96.39 & 95.11 & 96.48 & 96.39 & 98.13 & 95.19 & 88.84 & 99.03 & 95.62 & 96.91 & 89.53 & 97.08 & 99.01 & 96.22 \\
8 & 21 & 182 & 89.25 & 97.81 & 98.24 & 92.53 & 98.46 & 96.92 & 98.43 & 95.60 & 89.21 & 99.12 & 95.82 & 92.09 & 83.85 & 97.14 & 99.21 & 95.82 \\
9 & 32 & 282 & 87.94 & 87.30 & 87.73 & 87.52 & 88.58 & 90.92 & 97.97 & 91.38 & 100.00 & 99.35 & 88.51 & 90.92 & 99.65 & 88.72 & 99.52 & 88.23 \\
10 & 25 & 223 & 93.27 & 89.62 & 90.04 & 91.39 & 90.13 & 92.74 & 95.13 & 90.49 & 90.18 & 97.27 & 93.27 & 95.16 & 86.46 & 91.57 & 97.65 & 93.09 \\
11 & 31 & 274 & 86.50 & 93.00 & 93.43 & 91.09 & 94.82 & 91.90 & 95.13 & 94.60 & 94.38 & 95.75 & 90.15 & 81.24 & 88.47 & 91.24 & 95.89 & 89.93 \\
12 & 19 & 162 & 85.36 & 89.08 & 89.51 & 87.16 & 87.41 & 89.26 & 96.63 & 93.46 & 95.12 & 95.43 & 90.97 & 84.94 & 89.14 & 91.98 & 95.71 & 93.25 \\
13 & 27 & 241 & 88.53 & 90.28 & 90.71 & 88.63 & 94.77 & 93.86 & 97.13 & 88.63 & 87.88 & 98.56 & 90.85 & 92.45 & 94.44 & 92.95 & 98.62 & 91.79 \\
14 & 10 & 85 & 86.47 & 96.04 & 96.47 & 93.41 & 90.12 & 92.71 & 96.00 & 95.76 & 94.31 & 94.24 & 91.12 & 76.94 & 78.10 & 96.24 & 95.39 & 89.18 \\ \hline 
AA & \textbf{331} & \textbf{2917} & 88.26 & 91.40 & 92.13 & 90.41 & 92.80 & 92.90 & \textbf{96.48} & \textbf{92.97} & 92.82 & \textbf{97.51} & \textbf{93.18} & 89.20 & 87.91 & 93.03 & \textbf{97.75} & \textbf{93.25} \\
OA &  &  & 86.90 & 90.11 & 91.83 & 89.96 & 92.94 & 92.79 & \textbf{96.78} & \textbf{92.98} & 94.52 & \textbf{97.91} & \textbf{93.68} & 89.87 & 89.70 & 97.78 & \textbf{97.96} & \textbf{93.79} \\
$\kappa$ &  &  & 86.23 & 90.02 & 91.15 & 89.11 & 92.35 & 92.19 & \textbf{96.49} & \textbf{92.48} & 93.96 & \textbf{97.39} & \textbf{92.99} & 87.07 & 85.59 & 91.51 & \textbf{97.43} & \textbf{93.01} \\
Time &  &  &  & 56.75 & 3.16 & 0.14 & 12.30 & 3.18 & \textbf{0.43} & \textbf{0.38} & 7.54 & \textbf{9.32} & \textbf{9.1} & 1.50 & 2.26 & 5.28 & \textbf{9.39} & \textbf{9.18} \\ \hline 
\end{tabular}
}
\end{table*}




\subsection{Comparison with other DR methods} \label{sec:comparison}
Table~\ref{tab:res_IndP_data} and \ref{tab:res_Bots_data} provide the statistics of the class-wise, average, kappa and overall classification accuracy with the execution time of the algorithms for Indian pine and Botswana dataset respectively. The experiments were conducted by randomly chosen training samples and the results are averaged over five iterations for each case. %
From Table~\ref{tab:res_IndP_data} and \ref{tab:res_Bots_data}, we can observe that

\begin{itemize}

\item S3RMLSC with HGF filter yields the best classification performance among these supervised and semi-supervised DR methods on both the datasets.

\item The use of HGF filter significantly improves the classification accuracy of all methods. In Indian Pines dataset (Table~\ref{tab:res_IndP_data}), we observed a performance boost of 15\% with the use of HGF filter while the improvement in Botswana (Table~\ref{tab:res_Bots_data}) is around 4\%.

\item Among the spectral domain supervised DR methods, MLSC outperforms the other methods with a few exceptions. For example, in Botswana dataset, RLDE performs better than MLSC when HGF is not applied. However, the classification accuracy of MLSC is close to the best performing methods while its time complexity is very low.

\item The performance of NWFE is very competitive with MLSC without HGF condition in both the datasets. 
However, the time complexity of NWFE is 30 to 100 times more than the time complexity of MLSC. For real-time operation, the proposed MLSC might be a suitable solution with reliable performance.

\item MLSC extends the LSC along the nonlinear manifold and achieves better overall accuracy than LSC in both the datasets. Additionally, the computational complexity of MLSC is ten times smaller than that of LSC.


\item SSRMLSC outperforms the spectral base supervised, semi-supervised and other spatial-spectral DR methods. Though RLDE performs better than MLSC, the performance of its spatial extension SSRLDE is lower than SSRMLSC (considering the case when HGF is not used). 

\item The use of unlabeled data samples in semi-supervised learning can negatively affect the classification performance as in the case of SELDLPP and SELDNPE. 


\item From Table~\ref{tab:res_IndP_data}, we can observe that all state-of-the-art methods fail to classify for classes~$\{1,7,9,16\}$ properly due to insufficient data. However, the regularization terms in the proposed methods avoid this problem and classify these classes correctly. 

\item Comparing the computation time, S3RMLSC is comparatively slower than the other semi-supervised methods, but it provides faster performance in comparison to SSLSC method. Similarly, the time consumed by the MLSC and SSRMLSC are competitive with other state-of-the-art DR methods while achieving reliable performance.

\item The proposed supervised and semi-supervised methods also perform well for very small size labeled training samples. Table~\ref{tab:OA_small_sample_size} supports the effectiveness of the proposed methods. 

\item In this work, we performed most of the analysis using the Botswana and Indian pines data. However, to show the effectiveness of the parameters learned from these analysis, we tested it on the Pavia university dataset ($610 \times 340$ spatial pixels, $103$ spectral bands and $9$ classes).  
From Table~\ref{tab:OA_small_sample_size}, we can observe that the proposed algorithms boost the performance to a greater extent on Pavia university data in the presence of HGF filters. For instance,  $5-7\%$ improvement is observed with $8$ training samples, while the improvement is $8-12\%$ for $20$ training data samples. Similarly, the proposed algorithms also achieves a performance boost of $3-4\%$  and $4-6\%$ for $8$ and $20$ training data samples respectively while HGF is not used.

\end{itemize}

\begin{table}[!h]
\centering
\caption{Classification performance in the case of small sample size, i.e., 8, 10, 15, and 20 labeled samples per class.}
\label{tab:OA_small_sample_size}
\resizebox{0.47\textwidth}{!}{
\begin{tabular}{cc|cc|c|cc|c|cc}
\hline
\multicolumn{10}{c}{Indian Pines} \\ \hline
\multicolumn{1}{c|}{\begin{tabular}[c]{@{}c@{}}No. of labeled\\ train samples\end{tabular}} & \multicolumn{1}{c|}{\multirow{2}{*}{RLDE}} & \multicolumn{2}{c|}{MLSC} & \multicolumn{1}{c|}{\multirow{2}{*}{SSRLDE}} & \multicolumn{2}{c|}{SSRMLSC} & \multicolumn{1}{c|}{\multirow{2}{*}{SSLSC}} & \multicolumn{2}{c}{S3RMLSC} \\ \cline{3-4} \cline{6-7} \cline{9-10} 
\multicolumn{1}{l|}{} & \multicolumn{1}{c|}{} & \multicolumn{1}{c|}{HGF} & \multicolumn{1}{c|}{No HGF} & \multicolumn{1}{c|}{WMF} & \multicolumn{1}{c|}{HGF} & \multicolumn{1}{c|}{No HGF} & \multicolumn{1}{c|}{} & \multicolumn{1}{c|}{HGF} & \multicolumn{1}{c}{No HGF} \\ \hline %
\multicolumn{1}{c|}{8} & 53.21 & 71.04 & \multicolumn{1}{|c|}{53.70} & 58.70 & 76.01 & \multicolumn{1}{|c|}{58.20} & 48.72 & 75.89 & \multicolumn{1}{|c}{58.90} \\
\multicolumn{1}{c|}{10} & 55.06 & 74.22 & \multicolumn{1}{|c|}{55.67} & 69.99 & 78.70 & \multicolumn{1}{|c|}{55.26} & 49.51 & 78.77 & \multicolumn{1}{|c}{55.62} \\
\multicolumn{1}{c|}{15} & 61.91 & 81.57 & \multicolumn{1}{|c|}{61.69} & 73.95 & 84.40 & \multicolumn{1}{|c|}{61.97} & 56.89 & 84.60 & \multicolumn{1}{|c}{62.57} \\
\multicolumn{1}{c|}{20} & 63.84 & 82.70 & \multicolumn{1}{|c|}{63.89} & 76.16 & 85.50 & \multicolumn{1}{|c|}{64.27} & 59.32 & 85.59 & \multicolumn{1}{|c}{64.96} \\ \hline \hline
\multicolumn{10}{c}{Botswana} \\ \hline
\multicolumn{1}{c|}{\begin{tabular}[c]{@{}c@{}}No. of labeled\\ train samples\end{tabular}} & \multicolumn{1}{c|}{\multirow{2}{*}{RLDE}} & \multicolumn{2}{c|}{MLSC} & \multicolumn{1}{c|}{\multirow{2}{*}{SSRLDE}} & \multicolumn{2}{c|}{SSRMLSC} & \multicolumn{1}{c|}{\multirow{2}{*}{SSLSC}} & \multicolumn{2}{c}{S3RMLSC} \\ \cline{3-4} \cline{6-7} \cline{9-10} 
\multicolumn{1}{l|}{} & \multicolumn{1}{c|}{} & \multicolumn{1}{c|}{HGF} & \multicolumn{1}{c|}{ No HGF} & \multicolumn{1}{c|}{WMF} & \multicolumn{1}{c|}{HGF} & \multicolumn{1}{c|}{No HGF} & \multicolumn{1}{c|}{} & \multicolumn{1}{c|}{HGF} & \multicolumn{1}{c}{No HGF} \\ \hline %
\multicolumn{1}{c|}{8} & 85.66 & 89.87 & \multicolumn{1}{|c|}{86.22} & 85.96 & 90.64 & \multicolumn{1}{|c|}{86.58} & 84.86 & 90.75 & \multicolumn{1}{|c}{86.44} \\
\multicolumn{1}{c|}{10} & 86.98 & 90.33 & \multicolumn{1}{|c|}{87.80} & 87.18 & 92.23 & \multicolumn{1}{|c|}{88.47} & 86.56 & 92.55 & \multicolumn{1}{|c}{88.51} \\
\multicolumn{1}{c|}{15} & 90.17 & 94.17 & \multicolumn{1}{|c|}{90.31} & 92.85 & 95.09 & \multicolumn{1}{|c|}{90.87} & 89.59 & 95.12 & \multicolumn{1}{|c}{90.76} \\
\multicolumn{1}{c|}{20} & 91.33 & 95.90 & \multicolumn{1}{|c|}{91.52} & 93.77 & 97.19 & \multicolumn{1}{|c|}{91.67} & 91.04 & 97.27 & \multicolumn{1}{|c}{91.97} \\ \hline \hline
\multicolumn{10}{c}{Pavia University} \\ \hline
\multicolumn{1}{c|}{\begin{tabular}[c]{@{}c@{}}No. of labeled\\ train samples\end{tabular}} & \multicolumn{1}{c|}{\multirow{2}{*}{RLDE}} & \multicolumn{2}{c|}{MLSC} & \multicolumn{1}{c|}{\multirow{2}{*}{SSRLDE}} & \multicolumn{2}{c|}{SSRMLSC} & \multicolumn{1}{c|}{\multirow{2}{*}{SSLSC}} & \multicolumn{2}{c}{S3RMLSC} \\ \cline{3-4} \cline{6-7} \cline{9-10} 
\multicolumn{1}{l|}{} & \multicolumn{1}{c|}{} & \multicolumn{1}{c|}{HGF} & \multicolumn{1}{c|}{ No HGF} & \multicolumn{1}{c|}{WMF} & \multicolumn{1}{c|}{HGF} & \multicolumn{1}{c|}{No HGF} & \multicolumn{1}{c|}{} & \multicolumn{1}{c|}{HGF} & \multicolumn{1}{c}{No HGF} \\ \hline %
\multicolumn{1}{c|}{8} & 65.14 & 72.34 & \multicolumn{1}{|c|}{68.21} & 66.52 & 72.89 & \multicolumn{1}{|c|}{69.66} & 65.17 & 72.95 & \multicolumn{1}{|c}{69.94} \\
\multicolumn{1}{c|}{10} & 67.19 & 75.81 & \multicolumn{1}{|c|}{70.65} & 69.74 & 76.21 & \multicolumn{1}{|c|}{71.31} & 66.73 & 76.00 & \multicolumn{1}{|c}{71.36} \\
\multicolumn{1}{c|}{15} & 68.06 & 78.55 & \multicolumn{1}{|c|}{72.90} & 72.07 & 78.91 & \multicolumn{1}{|c|}{73.72} & 65.84 & 78.75 & \multicolumn{1}{|c}{73.66} \\
\multicolumn{1}{c|}{20} & 70.16 & 80.34 & \multicolumn{1}{|c|}{74.93} & 73.96 & 80.81 & \multicolumn{1}{|c|}{76.73} & 66.57 & 80.88 & \multicolumn{1}{|c}{76.97} \\ \hline
\end{tabular}}
\end{table}

\subsection{Effect of Number of Reduced Dimensions}

\begin{figure}[!b]
	\centering
	\begin{subfigure}{.248\textwidth}
		\centering
		\includegraphics[width=.99\linewidth]{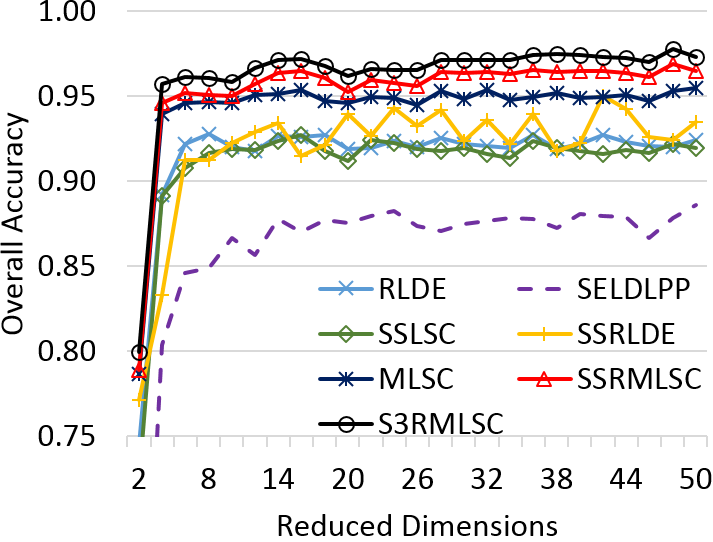}
		\caption{}
		\label{fig:All_Bots_10perc}
	\end{subfigure}%
    \begin{subfigure}{.248\textwidth}
		\centering
		\includegraphics[width=.99\linewidth]{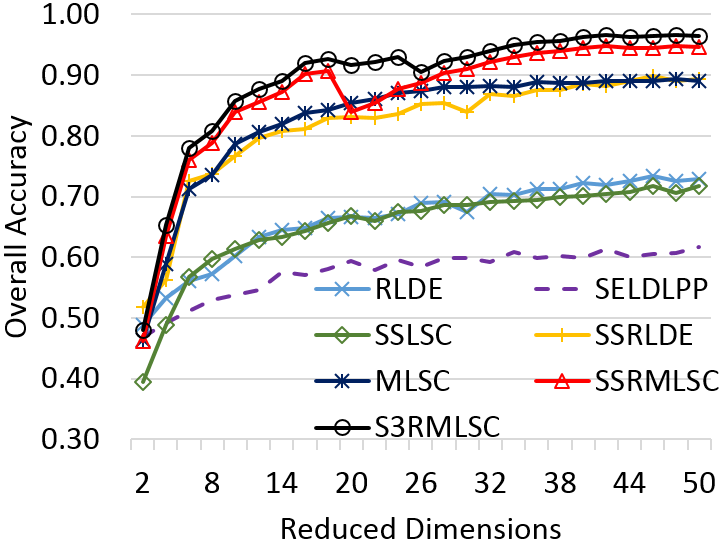}
		\caption{}
		\label{fig:All_IndP_10perc}
	\end{subfigure}
	\caption{ OA of different methods with increasing of subspace dimensions (\ref{fig:All_Bots_10perc}) Botswana and (\ref{fig:All_IndP_10perc}) Indian pine.}
	\label{fig:OA_multiDimens}
\end{figure}

Fig.~\ref{fig:OA_multiDimens} shows the variations of the overall accuracy with respect to different reduced dimensions for supervised and semi-supervised methods. We can observe that S3RMLSC, SSRMLSC and MLSC methods outperform the other state-of-art DR methods at various reduced dimensions. SSRLDE is the close competitor of the proposed semi-supervised method. However, its performance is lower than MLSC in Botswana dataset. S3RMLSC achieves significant accuracy at $12-15$ dimensions in  Indian pine and $8-10$ dimensions in Botswana  dataset. However, it is observed that its performance gradually becomes consistent 
afterward due to the redundancy of spectral bands. The performance in Fig.~\ref{fig:OA_multiDimens} proves the robustness of the proposed algorithm on different dimensions. 

\begin{figure}[!b]
	\centering
	\begin{subfigure}{.249\textwidth}
		\centering
		\includegraphics[width=.99\linewidth]{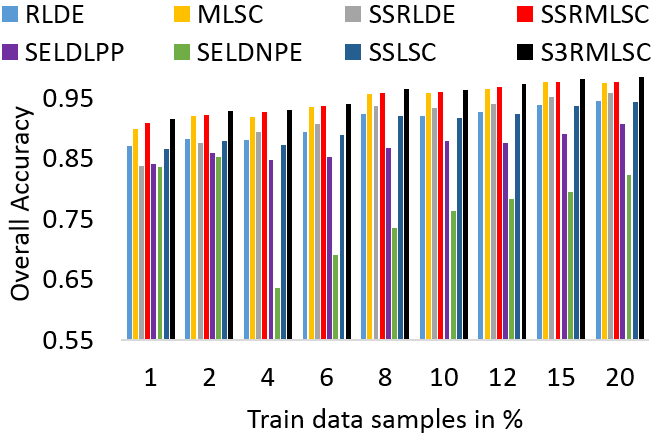}
		\caption{}
		\label{fig:All_Bots_30Dims}
	\end{subfigure}%
	\begin{subfigure}{.249\textwidth}
		\centering
		\includegraphics[width=.99\linewidth]{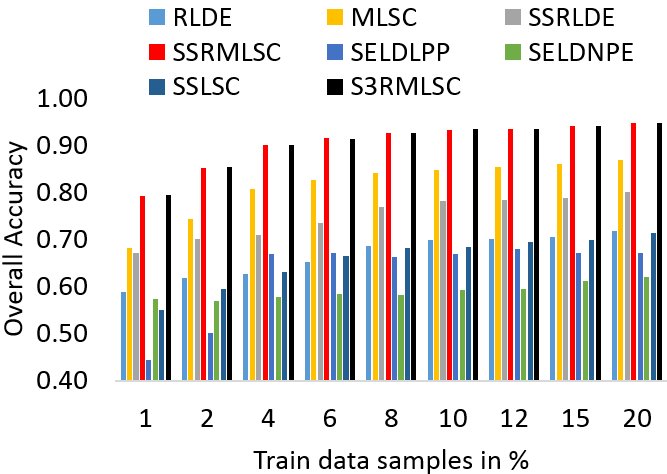}
		\caption{}
		\label{fig:All_IndP_30Dims}
	\end{subfigure}
	\caption{ OA of different methods with increase in labeled training data samples (in \%) per class (\ref{fig:All_Bots_30Dims}) Botswana and (\ref{fig:All_IndP_30Dims}) Indian pine.} 
	\label{fig:OA_Multiple_Data}
\end{figure}
\subsection{Effect of Number of Train Samples}

\begin{figure}[t]
	\centering
    \begin{subfigure}{.12\textwidth}
		\centering
		\includegraphics[width=.95\linewidth]{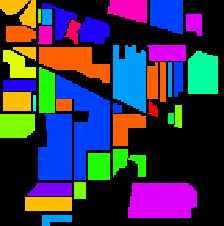}
		\caption{}
		\label{fig:IndP_GT}
	\end{subfigure}%
    \begin{subfigure}{.12\textwidth}
		\centering
		\includegraphics[width=.95\linewidth]{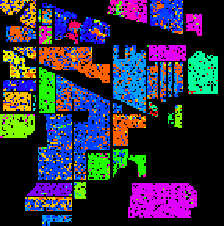}
		\caption{}
		\label{fig:IndP_RLDE}
	\end{subfigure}%
	\begin{subfigure}{.12\textwidth}
		\centering
		\includegraphics[width=.95\linewidth]{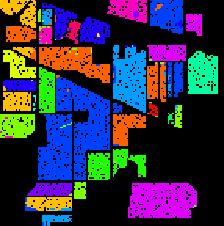}
		\caption{}
		\label{fig:IndP_MLSC}
	\end{subfigure} 
    \begin{subfigure}{.12\textwidth}
		\centering
		\includegraphics[width=.95\linewidth]{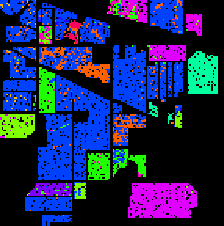}
		\caption{}
		\label{fig:IndP_SSRLDE}
	\end{subfigure}
    \begin{subfigure}{.12\textwidth}
		\centering
		\includegraphics[width=.95\linewidth]{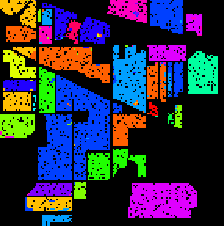}
		\caption{}
		\label{fig:IndP_SSRMLSC}
	\end{subfigure}%
    \begin{subfigure}{.12\textwidth}
		\centering
		\includegraphics[width=.95\linewidth]{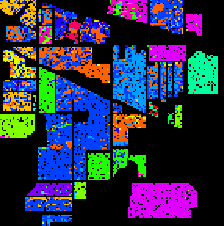}
		\caption{}
		\label{fig:IndP_SSLSC}
	\end{subfigure}%
	\begin{subfigure}{.12\textwidth}
		\centering
		\includegraphics[width=.95\linewidth]{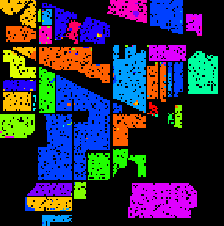}
		\caption{}
		\label{fig:IndP_S3RMLSC}
	\end{subfigure}
    \caption{ Classification maps for Indian pines using $10\%$ labeled and $2000$ unlabeled training pixels with 50 dimensions. (\ref{fig:IndP_GT}) Ground Truth, (\ref{fig:IndP_RLDE}) RLDE [OA=$72.64$], (\ref{fig:IndP_MLSC}) MLSC [OA=$95.38$], (\ref{fig:IndP_SSRLDE}) SSRLDE [OA=$77.13$], (\ref{fig:IndP_SSRMLSC}) SSRMLSC [OA = $96.36$], (\ref{fig:IndP_SSLSC}) SSLSC [OA=$71.72$], (\ref{fig:IndP_S3RMLSC}) S3RMLSC [OA=$96.46$].}
	\label{fig:OA_IndP_All}
\end{figure}

\begin{figure}[t]
	\centering
     \begin{subfigure}{.08\textwidth}
		\centering
		\includegraphics[width=.98\linewidth]{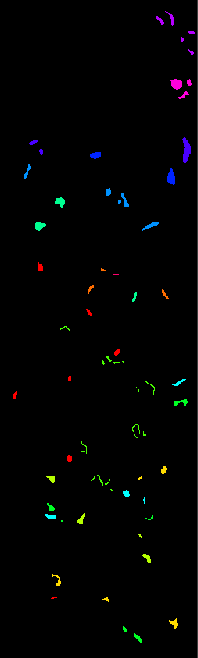}
		\caption{}
		\label{fig:Bots_GT}
	\end{subfigure}%
    \begin{subfigure}{.08\textwidth}
		\centering
		\includegraphics[width=.98\linewidth]{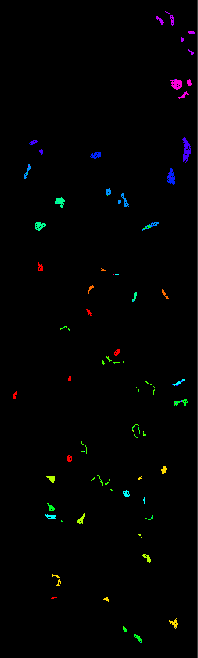}
		\caption{}
		\label{fig:Bots_RLDE}
	\end{subfigure}%
	\begin{subfigure}{.08\textwidth}
		\centering
		\includegraphics[width=.98\linewidth]{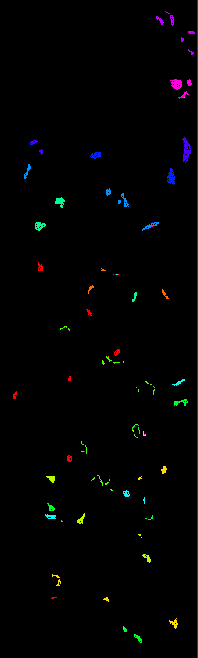}
		\caption{}
		\label{fig:Bots_MLSC}
	\end{subfigure}
    \begin{subfigure}{.08\textwidth}
		\centering
		\includegraphics[width=.98\linewidth]{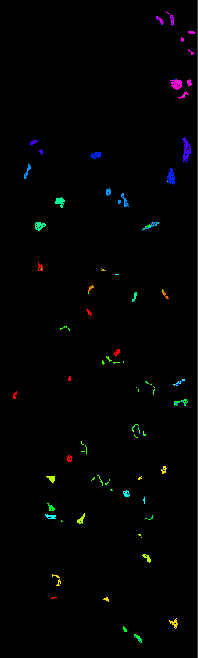}
		\caption{}
		\label{fig:Bots_SSRLDE}
	\end{subfigure}   
    \begin{subfigure}{.08\textwidth}
		\centering
		\includegraphics[width=.98\linewidth]{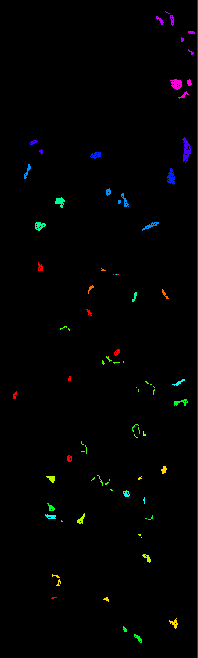}
		\caption{}
		\label{fig:Bots_SSRMLSC}
	\end{subfigure}%
    \begin{subfigure}{.08\textwidth}
		\centering
		\includegraphics[width=.98\linewidth]{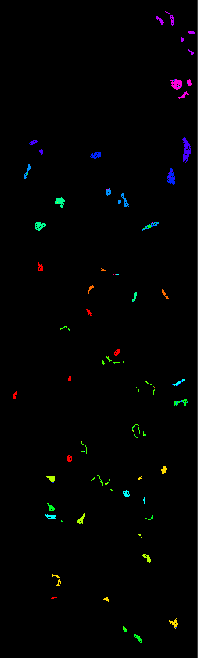}
		\caption{}
		\label{fig:Bots_SSLSC}
	\end{subfigure}%
	\begin{subfigure}{.08\textwidth}
		\centering
		\includegraphics[width=.98\linewidth]{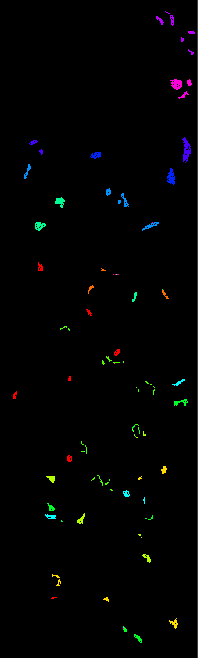}
		\caption{}
		\label{fig:Bots_S3RMLSC}
	\end{subfigure}
    \caption{ Classification maps for Botswana using $10\%$ labeled and $2000$ unlabeled training pixels with 50 dimensions. (\ref{fig:Bots_GT}) Ground Truth, (\ref{fig:Bots_RLDE}) RLDE [OA=$91.45$], (\ref{fig:Bots_MLSC}) MLSC [OA= $96.19$], (\ref{fig:Bots_SSRLDE}) SSRLDE [OA=$92.21$], (\ref{fig:Bots_SSRMLSC}) SSRMLSC [OA = $97.02$], (\ref{fig:Bots_SSLSC}) SSLSC [OA=$91.98$], (\ref{fig:Bots_S3RMLSC}) S3RMLSC [OA=$97.74$].}
	\label{fig:OA_Bots_All}
\end{figure}

Fig.~\ref{fig:OA_Multiple_Data} shows the effect of the number of labeled train data on classification performance. 
As per the experimental observations on Botswana (Figs.~\ref{fig:All_Bots_30Dims}) and Indian pine (\ref{fig:All_IndP_30Dims}), the proposed method significantly outperforms the other state-of-the-art methods. Table~\ref{tab:OA_small_sample_size} also proves the effectiveness of the proposed methods on small size training samples like $[8,10,15,20]$ samples per class. 
Fig.~\ref{fig:OA_IndP_All} and \ref{fig:OA_Bots_All} show the classification maps of the Indian pine and Botswana images for different methods in a single run. 

From these above-observed results, we can highlight the following conclusions
\begin{enumerate}[label= \roman*)]
\item A large number of the bands in an HSI data are redundant, and the intrinsic relevant information lies in a few intrinsic dimensions. Hence, DR improves the HSI classification performance by projecting the data to a reduced feature space where the effects of redundant bands are lessened.
\item MLSC performs better than supervised graph cut methods (SC and LSC), whereas SSRMLSC performs way better than SSRLDE. Further, S3RMLSC achieves significant performance improvement compared to the other supervised as well as semi-supervised methods. 
\item The proposed S3RMLSC method not only gives the best overall accuracy with highest class wise average performance but also gives the best kappa coefficient compared to others using limited bands in an optimal time. Hence, this explains the effectiveness of the algorithm in terms of optimal space and performance. 
\item Most of the state-of-the-art methods compromise the performance gain with the computational complexity. However, the time complexity of S3RMLSC is competitive with other methods while achieving promising performances. 
\item The proposed method also outperforms the other state-of-the-art methods when training set size is small.
\item S3RMLSC gives the best classification performance for both Botswana and Indian pine data with the same set of optimal parameters. Whereas, other methods tune their parameters based on the dataset. This demonstrates the robustness of the proposed algorithm over others.
\item The robustness of the proposed methods are also demonstrated by learning the parameters from the Indian pines and Botswana data and successfully using them on Pavia university data to obtain better classification results. 
\item The use of labeled and unlabeled data samples in the proposed semi-supervised DR method exploits the local property as well as the global geometry of the data. This gives the performance edge to S3RMLSC over MLSC.
\end{enumerate}



\section{Conclusion} \label{sec:conclusn}
In this paper, we propose S3RMLSC which uses both the spectral and spatial information to maximize the class discrimination. Spectral RMLSC incorporates the spectral information with a regularization term which overcomes the the data singularity by diversifying the HSI data samples. 
This enhances the discrimination capability and improves the classification accuracy. The NPMLSC method is a robust graph cut based spatial segmentation technique, which incorporates the spectral neighborhood measure with the spatial pixel neighborhood correlation to improve the class dissimilarity matrices. S3RMLSC takes the advantage of both spectral RMLSC and NPMLSC to obtain the optimal projection direction.  
The idea of maximizing the local patch margin from dissimilar classes while maintaining the individual patch compactness of manifold, makes our method theoretically and practically appealing. The selection of data samples from the patch-wise locality of the manifold retains the geometrical and nonlinear property of the data.  Apart from that, the use of HGF increases the neighboring pixel consistency, preserves the spatial contextual information and discriminates the edges of the complimentary information robustly. We tested our method and other classical methods on two popular real-world HSI datasets.  On these experiments, the proposed method consistently outperforms the classical methods by a large margin. These promising experimental results of S3RMLSC on different datasets demonstrate its robustness as well as generic applicability.
Further, we aim to explore the multivariate tensorial extension of this method in our future studies.


 \bibliographystyle{IEEEtran}
 \bibliography{trans_14_12_16}

\end{document}